\documentclass[letterpaper, 10 pt, journal, twoside]{IEEEtran} 

\usepackage{graphics}
\usepackage{epsfig}
\usepackage{mathptmx}
\usepackage{times}
\usepackage{amsmath}
\usepackage{amssymb}
\usepackage{booktabs} 
\usepackage{cite}
\usepackage{array}
\usepackage{amsfonts}
\usepackage{amsbsy}
\usepackage{placeins}
\usepackage{multirow}
\usepackage{tabularx}
\usepackage{algorithm}
\usepackage{algpseudocode}
\usepackage{xcolor}
\usepackage{url}
\usepackage{hyperref}

\begin{document}

\title{Trajectory Tracking Control of Dual-PAM Soft Actuator with Hysteresis Compensator}

\newcommand{\acceptedNote}{This paper has been published in the IEEE Robotics and Automation Letters \href{https://doi.org/10.1109/LRA.2023.3334098}{DOI 10.1109/LRA.2023.3334098}, copyright has been transferred to the IEEE. Final version is available at IEEE Xplore: \url{https://ieeexplore.ieee.org/document/10321653}.}

\author{Junyi Shen$^{1}$, Tetsuro Miyazaki$^{1}$, Shingo Ohno$^{2}$, Maina Sogabe$^{1}$, and Kenji Kawashima$^{1}$%
\thanks{Manuscript received: August 6, 2023; Revised: October 10,2023; Accepted: November 8, 2023.}
\thanks{This paper was recommended for publication by Editor Yong-Lae Park upon evaluation of the Associate Editor and Reviewers' comments. This work was supported by Japan Society for the Promotion of Science (JSPS) under the grant KAKENHI 21H04544 and Bridgestone Corporation. \textit{(Corresponding author: Kenji Kawashima)}} 
\thanks{$^{1}$Junyi Shen, Tetsuro Miyazaki, Maina Sogabe, and Kenji Kawashima are with Department of Information Physics and Computing, The University of Tokyo, 113-8654, Tokyo, Japan (e-mail: \{junyi-shen, fsogabe-vet\}@g.ecc.u-tokyo.ac.jp, \{Tetsuro\_Miyazaki, kenji\_kawashima\}@ipc.i.u-tokyo.ac.jp).}%
\thanks{$^{2}$Shingo Ohno is with Innovative Project Planning and Promotion Department, Bridgestone Corporation, 104-0031, Tokyo, Japan (e-mail: shingo.oono@bridgestone.com).}%
\thanks{Digital Object Identifier (DOI): see top of this page.}
}

\markboth{IEEE Robotics and Automation Letters. Preprint Version. Accepted November, 2023}
{Shen \MakeLowercase{\textit{et al.}}: Trajectory Tracking Control of Dual-PAM Soft Actuator with Hysteresis Compensator} 

\maketitle

\textit{\acceptedNote} \\

\begin{abstract}

Soft robotics is a swiftly evolving field. Pneumatic actuators are suitable for driving soft robots because of their superior performance. However, their control is challenging due to the hysteresis characteristics. In response to this challenge, we propose an adaptive control method to compensate for the hysteresis of soft actuators. Employing a novel dual pneumatic artificial muscle (PAM) bending actuator, the innovative control approach abates hysteresis effects by dynamically modulating gains within a traditional PID controller corresponding to the predicted variation of the reference trajectory. Through experimental evaluation, we found that the proposed control method outperforms its conventional counterparts regarding tracking accuracy and response speed. Our work reveals a new direction for advancing model-free control in soft actuators.

\end{abstract}


\section{Introduction}

\IEEEPARstart{S}{oft} robotics, a field driven by flexible, lightweight structures with infinite degrees of freedom, offers a safe and natural human-machine interface during collisions or impacts \cite{rus2015design}. This field has seen exceptional growth due to the development of soft actuators constructed from resilient materials, such as silicone rubber and shape memory alloys \cite{AZAMI2019111623, huang2018chasing}. These materials' innate flexibility provides soft robots the adaptability for diverse applications, ranging from medical devices and rehabilitation to industrial tasks \cite{8700373, app9142869, doi:10.1089/soro.2013.0007}.

Soft robotic motion depends on deformations of the flexible actuators, with pneumatic actuation playing a significant role. Pneumatic artificial muscles (PAMs), based on the McKibben structure that replicates biological muscle movements, have been praised for their cost-effectiveness and high force-to-weight ratio \cite{aschemann2013comparison}. Pneumatic bending actuators, typified by multiple miniature compartments, further advance this field, enabling more versatile movements and interactions \cite{helps2018proprioceptive,zhao2015scalable,laschi2014soft}.

While the flexibility benefits soft robots, it also introduces challenges, particularly in tasks requiring precision and rapid response \cite{7487695}. A significant hurdle is hysteresis that makes the system complex to model and results in delayed actuator responses towards control signal changes \cite{10.1111/j.1096-3642.1985.tb01178.x}. This issue is further augmented in pneumatic actuators due to the compressibility, limiting the application of pneumatic soft robots \cite{helps2018proprioceptive}.

In addressing hysteresis, recent research has primarily embraced model-based control strategies and feedforward hysteresis compensation techniques \cite{zang2017position}. The former conceptualizes hysteresis as an unmodeled dynamic or disturbance, aiming to attenuate its effects by complex nonlinear control algorithms \cite{schreiber2011tracking}. Conversely, the latter leverages inversed hysteresis models derived through the amalgamation of assorted mathematical play operators \cite{xie2018hysteresis}. However, model-based nonlinear algorithms hinge on accurate linear approximation, presenting a limitation. Meanwhile, feedforward hysteresis compensations demand substantial computational effort, both online for inversion processes and offline for hysteresis model construction, and encounter challenges in mimicking asymmetrical hysteretic traits \cite{SHI201376, zhao2006neural, 6648378}. In addition, the contemporary state-of-the-art primarily targets simple soft actuators possessing a singular degree of freedom (DoF) or multi-DoF actuators that can be segregated into elementary 1-DoF actuation units. Such prevalent methodologies may stumble when confronted with actuators exhibiting indivisible complex structures. Given the intrinsic complexities, potential unreliabilities of current methods, and their inadequacies for high-DoF actuators, the necessity for a straightforward yet efficacious anti-hysteresis control approach is accentuated \cite{rus2015design, laschi2014soft}. 

This work presents precise bending trajectory tracking control on a novel dual-PAM bending actuator. The architecture of this actuator features two antagonistically arranged PAMs. This arrangement melds the compact structure inherent in antagonistic mechanisms with the potentially adjustable stiffness and notable power-to-volume ratio provided by PAMs \cite{ZHU20082248}. Unlike traditional parallel manipulators, whose motions can be dissected into simple linear motions of individual units \cite{6648378, xie2020trajectory}, our dual-PAM actuator facilitates bending actuation via the correlated continuum deformation of both PAMs alongside the elastic metal plate. This driving mechanism resists decomposition and straightforward modeling. In response to this challenge, we introduce a novel scheme termed \textit{adaptforward} control. This strategy, rooted in the simple, effective, and ubiquitously employed PID controller, dynamically tunes the feedback gains based on the anticipated variation of the reference signal, thereby attenuating the hysteresis affects in actuator manipulation. Our proposal is validated through comparative motion tracking experiments.

The discourse unfolds as follows: Section 2 elucidates the dual-PAM soft actuator's design, mechanism, and hysteresis traits. Section 3 unveils the new control strategy. Section 4 corroborates its effectiveness through experimental validation, juxtaposing its bending motion tracking accuracy against conventional methods. Section 5 encapsulates the conclusions.

\section{Dual-PAM Soft Bending Actuator} 

\subsection{Overall Configuration}

\begin{figure}[tb]
    \centering
\includegraphics[width=\linewidth]{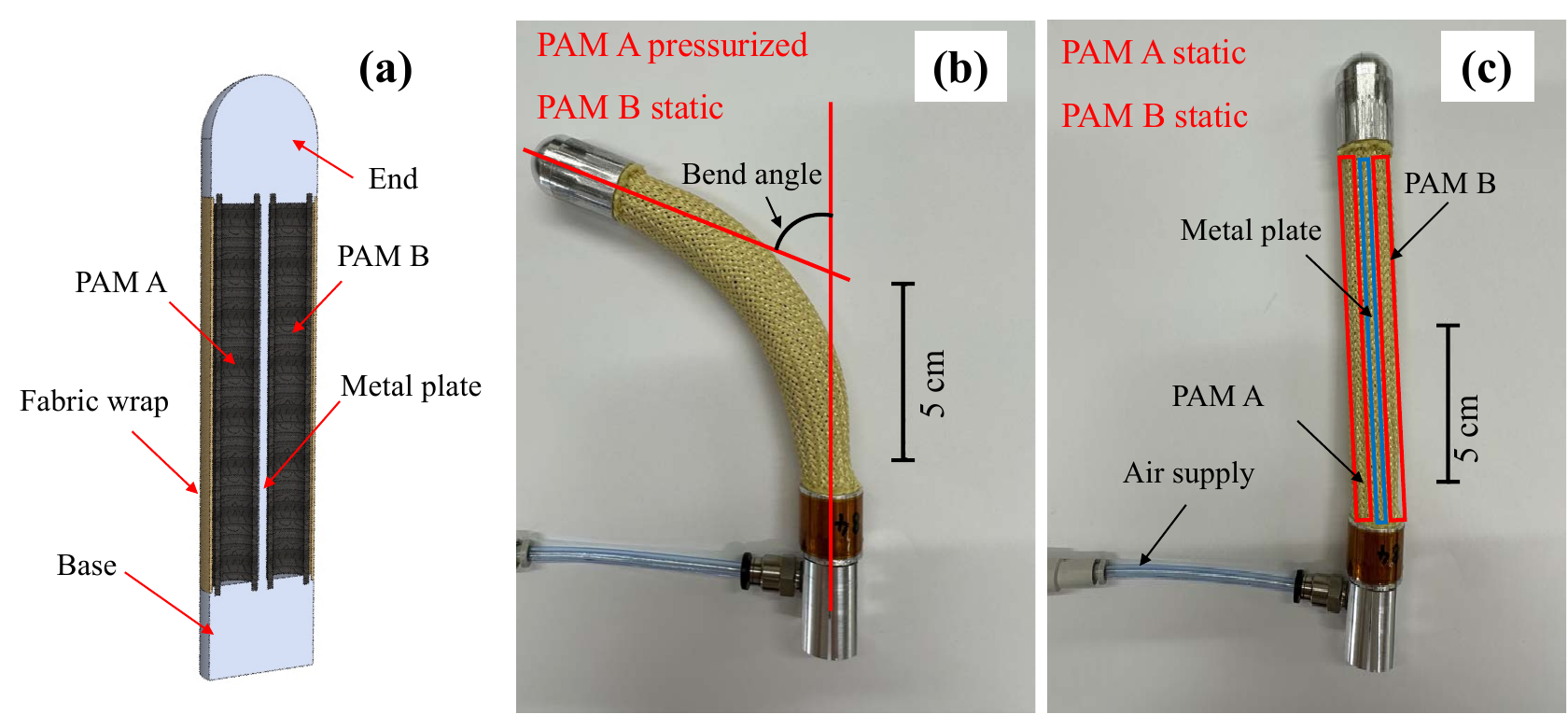}
    \caption{Dual-PAM bending actuator: (a) longitudinal section, (b) single PAM being pressurized, (c) both PAMs at static state.}
    \label{fig:proposed PAM}
\end{figure}

Fig. \ref{fig:proposed PAM} delineates the dual-PAM bending actuator employed in this study. The actuator's design encompasses a metallic base, a pair of rubber chambers, a metallic end, an elastic metal plate bridging the end and base, and a fabric sheath laterally enveloping the entire structure. This setup allows each chamber to operate as an independently controlled PAM; however, given their shared fabric sheath and common connection to the end and base, the actuation of each PAM is interdependent, resulting in an indivisible motion of the actuator and introducing specific modeling intricacies.

\begin{figure}[tb]
    \centering
\includegraphics[width=\linewidth]{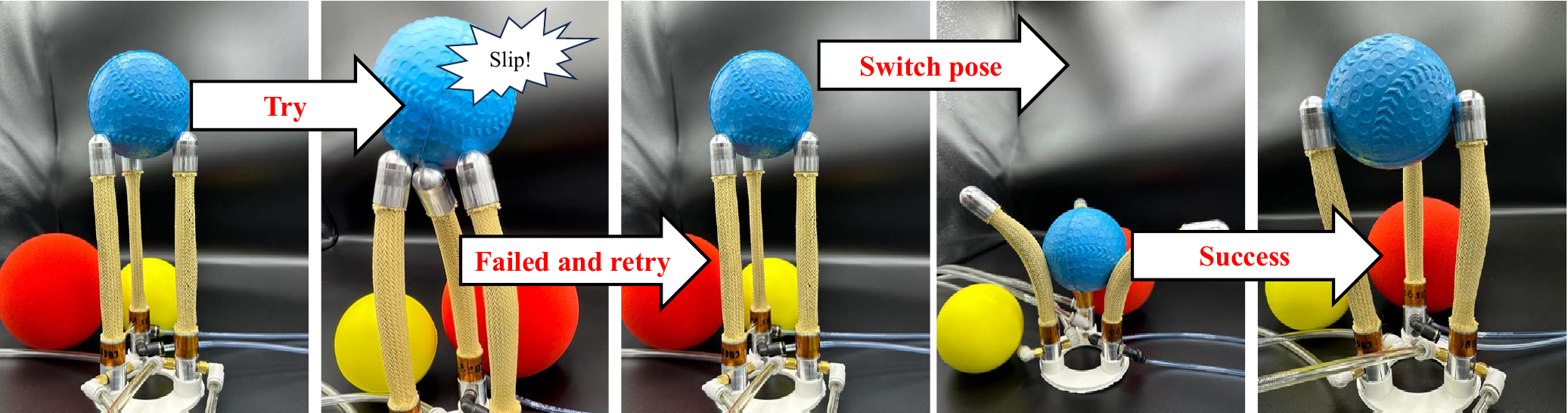}
    \caption{Application of the dual-PAM actuator in soft grippers that can change poses for grasping objects with unknown structures.}
    \label{fig:demo}
\end{figure}

The actuator achieves controlled bending by the mechanism of PAMs. As illustrated in Fig. \ref{fig:proposed PAM}(b), when one single PAM is pressurized, it contracts and generates an axial tensile force toward the end. By creating an inflation differential between two PAMs, an imbalance in bending torques is established on either side of the metal plate, instigating its folding and propelling the actuator's overall motion. The antagonistic setup of the two PAMs facilitates swift responsiveness to control signals and the bidirectional bending capability, augmenting its adaptability in unknown environments as illustrated in Fig. \ref{fig:demo}. Moreover, this configuration permits adjustable stiffness of the actuator through varying pressurization combinations.

The actuator's bending angle can be modulated by altering the internal pressures within each PAM. In a fully depressurized state, the actuator swiftly reverts to its original straight configuration, as exhibited in Fig. \ref{fig:proposed PAM} (c). This recovery is ascribed to the elasticity of the metal plate, provided the applied torque remains within its elastic limit, averting irreversible deformation and enabling natural reversion to the initial form.

Given the symmetrical bending performance rooting from the structural design of the actuator, our motion tracking analysis is confined to unidirectional bending in this work to reasonably minimize the investigative workload.

\subsection{Hysteresis Characteristics}

\subsubsection{Experimental Apparatus}

\begin{figure}[tb]
    \centering
\includegraphics[width=0.8\linewidth]{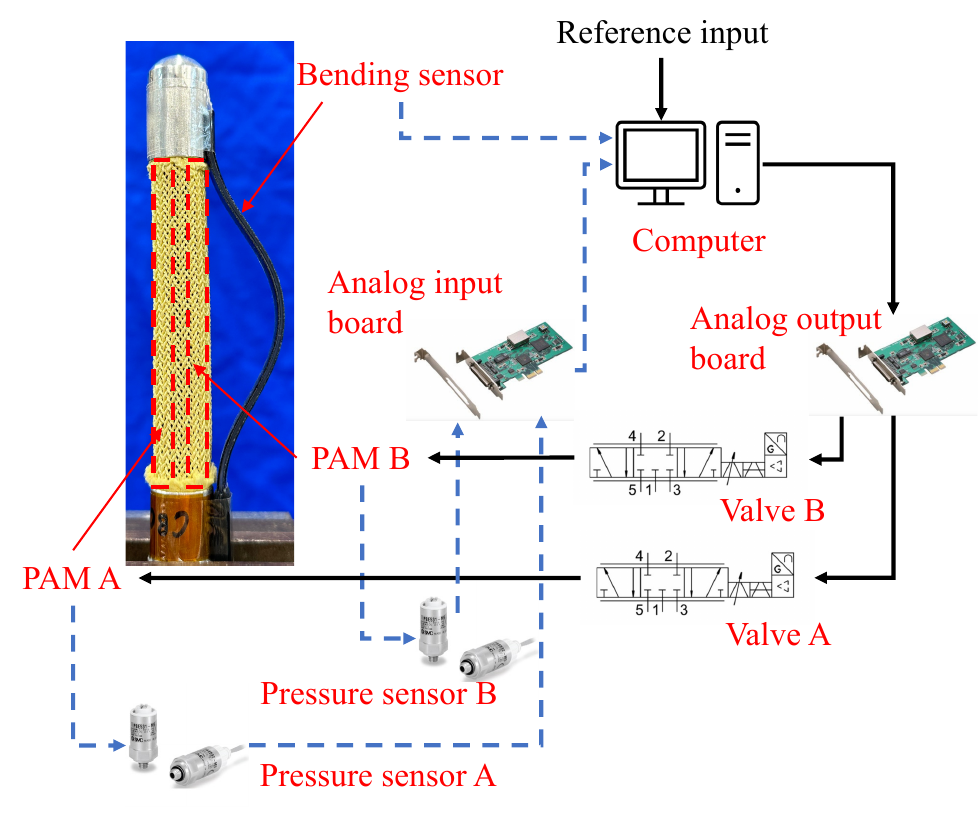}
    \caption{Experimental apparatus.}
    \label{fig:exp app}
\end{figure}

We obtained the bending-pressure hysteresis loops of the soft actuator by experiments. Experimental configuration is depicted in Fig. \ref{fig:exp app}, encompassing components include the dual-PAM actuator, a prototype provided by Bridgestone, a pair of Festo MPYE-5-M5-010B proportional valves, one BENDLABS-1AXIS angular sensor, two SMC PSE 540 A-R06 pressure sensors, and a computer system, within which a Contec AI-1616L-LPE and a Contec AO-1608L-LPE function respectively as the analog input board and the analog output board.

\subsubsection{Assessment of Hysteresis Loops}

\begin{figure}[tb]
    \centering
\includegraphics[width=\linewidth]{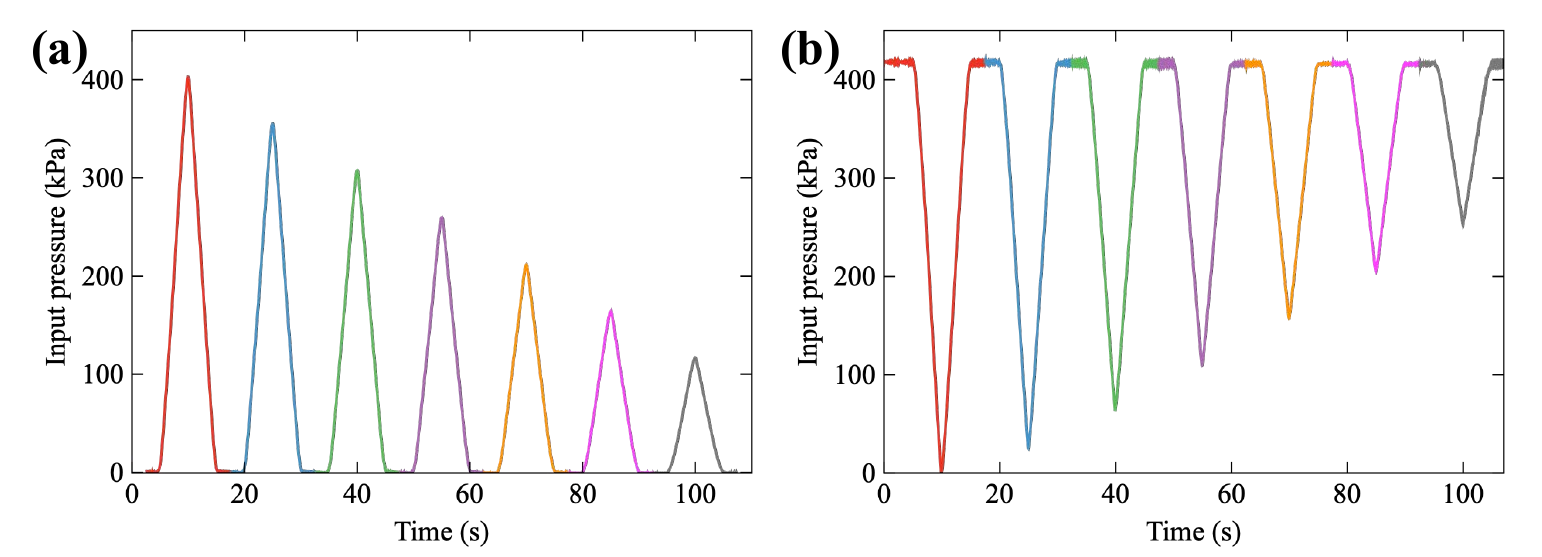}
    \caption{Pressure variations in the active PAM: (a) pressurization from an atmospheric value, (2) depressurization from a pre-pressurized state.}
    \label{fig:pre vari}
\end{figure}

Previous investigations into the hysteresis characteristics of PAMs employed triangular pressure waveforms originating from atmospheric states \cite{xie2018hysteresis}, thereby overlooking the hysteresis properties during transitions between depressurization and pressurization states at various deformation levels. To render a more comprehensive analysis of hysteresis properties, we design our assessment to encompass measurements of hysteresis loops emanating from two distinct initial states. In the first case, the internal pressures of both PAMs are set at atmospheric levels, with the initial bending angle defined as zero. Following this, a positive triangular waveform with diminishing amplitude alternately pressurizes and depressurizes PAM A, as demonstrated in Fig. \ref{fig:pre vari} (a). The second scenario posits an initial pressure of 420 kPa within PAM A, with a subsequent application of a negative triangular waveform with diminishing amplitude to alternately depressurize and pressurize it, as depicted in Fig. \ref{fig:pre vari} (b). The recorded minimum bending angle is designated as the zero-bending point in this latter measurement. Color distinctions within (a) and (b) delineate different input pressure cycles. Throughout the experimentation, continuous recordings of the actuator's bending angle and PAM A's internal pressure are maintained, while the internal pressure of PAM B is steadfastly held at the atmospheric value.

\begin{figure}[tb]
    \centering
\includegraphics[width=\linewidth]{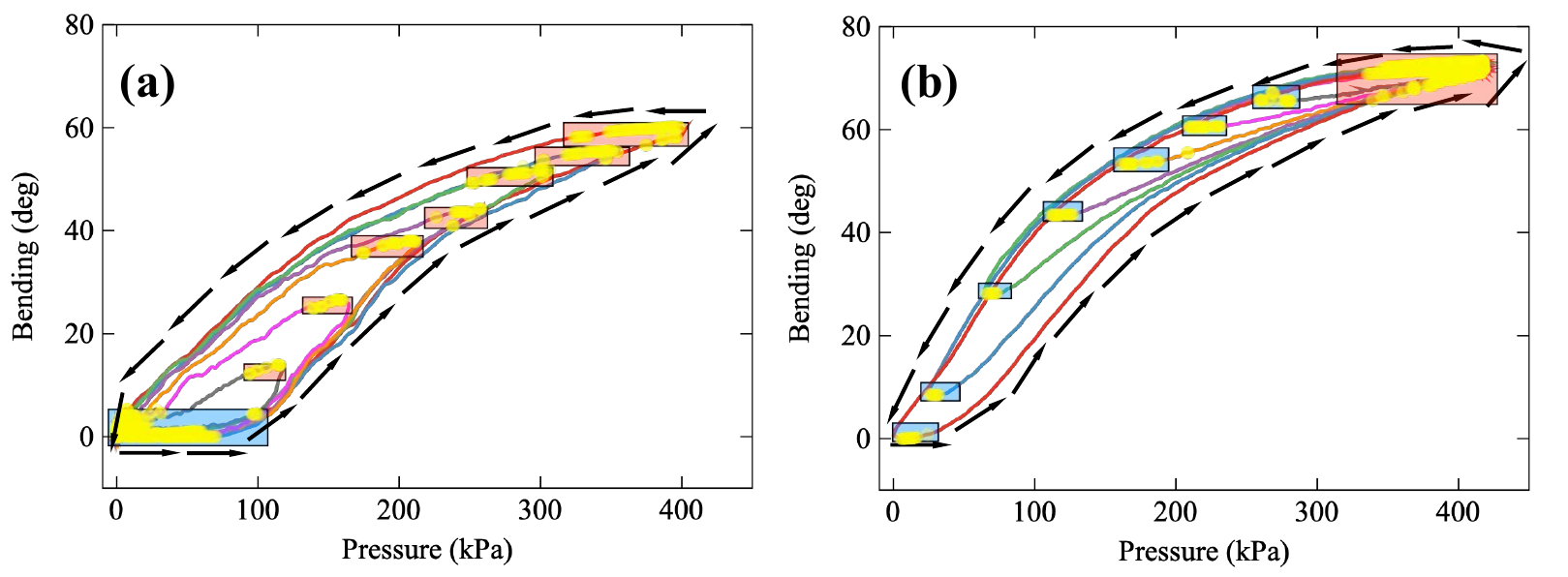}
    \caption{Hysteresis loops: (a) results from Fig. \ref{fig:pre vari} (a), (b): results from Fig. \ref{fig:pre vari} (b). Golden shadows indicate points within hysteresis dead-zones.}
    \label{fig:hysteresis}
\end{figure}

Resultant hysteresis loops, corresponding to the different pressure variations, are shown in Fig. \ref{fig:hysteresis} (a) and (b), respectively. Arrows indicate the directions, with color variations aligning with the different pressure circles depicted in Fig. \ref{fig:pre vari}.

To analyze the hysteresis dead-zones and quasi-dead-zones within each loop, we delineate specific detection criteria. Initially, a positive-definite average angle-pressure gradient is computed from their respective variation range. The gradients between the angle and pressure of sample points located at 30 percent proximal to both ends of the pressure range are computed, with their absolute values compared with the calculated average gradient. Any gradient falling below 30 percent of the average gradient is identified as within the dead-zones or quasi-dead-zones. The process is succinctly delineated in Algorithm \ref{alg:deadzone_detection}, where `\textit{gradientThreshold}' and `\textit{pressureThreshold}' are set to 0.3 and 0.2 respectively, conforming to our detection scheme. Pointed within dead-zone are accentuated by golden shadows in Fig. \ref{fig:hysteresis}. Dead-zones manifesting during the state transition from pressurization to depressurization are denoted in red, while those during the depressurization to pressurization transition are masked in blue. 

\begin{algorithm}
\caption{Detection of Deadzone in Hysteresis Loops}
\label{alg:deadzone_detection}
\begin{algorithmic}[1]
\Require
\Statex \textit{gradientThreshold}, \textit{pressureThreshold}
\Statex \textit{hysteresisLoops}, \textit{samples}
\Ensure
\Statex Identification of samples in deadzone

\State \textit{gradThreshold} \( \leftarrow \) \textit{gradientThreshold}
\State \textit{pressThreshold} \( \leftarrow \) \textit{pressureThreshold}

\For{each \textit{loop} in \textit{hysteresisLoops}}
    \State Compute average pressure: \( P_{\mathrm{ave}} \leftarrow (P_{\mathrm{max}} + P_{\mathrm{min}})/2 \)
    \State Compute positive-definite average gradient: \( \mathrm{grad}_{\mathrm{ave}} \leftarrow (\theta_{\mathrm{max}} - \theta_{\mathrm{min}})/(P_{\mathrm{max}} - P_{\mathrm{min}}) \)
    
    \For{each \textit{sample} in \textit{samples}}
        \If{ \( |\mathrm{grad}_{sample}| < \textit{gradThreshold} \times \mathrm{grad}_{\mathrm{ave}} \) \\ \textbf{and} \( |P - P_{\mathrm{ave}}|/(P_{\mathrm{max}} - P_{\mathrm{min}}) > \textit{pressThreshold} \) }
            \State Mark \textit{sample} as in dead-zone
        \Else
            \State Mark \textit{sample} as not in dead-zone
        \EndIf
    \EndFor
\EndFor
\end{algorithmic}
\end{algorithm}

An inspection of Fig. \ref{fig:hysteresis} (a) reveals an escalation in the widths of the red zones with increased bending angles. Conversely, Fig. \ref{fig:hysteresis} (b) exhibits an inverse correlation where the widths of the blue boxes amplify correspondingly with the diminution of bending angles. At highly pressurized states of the actuator, a noticeable increment in the width of blue boxes is captured with the escalation of bending angles, attributed to most part of the the hysteresis loop residing within the dead-zone as the actuator is highly deformed. The broadening of hysteresis dead-zones signifies a magnification in the actuator's hysteretic properties, aligning with \cite{chou1996measurement} that posits significant nonlinearities in PAMs when the actuator operates at extremely high or low pressurized levels.

\section{Trajectory Tracking Control Strategy}

Configuration of the dual-PAM actuator introduces a high-DoF nature into the control system, significantly complicating the application of model-based control algorithms or feedforward hysteresis compensators. In this work, we deploy a model free feedback control system comprising two separated subsystems with cascaded loops and newly proposed adaptive hysteresis compensators. By adhering to only fundamental guidelines, our system design reduces efforts for alleviating hysteresis effects in precisely controlling soft actuators with complex structures.

\subsection{Design of Control System}

\begin{figure*}[tb]
    \centering
\includegraphics[width=\linewidth]{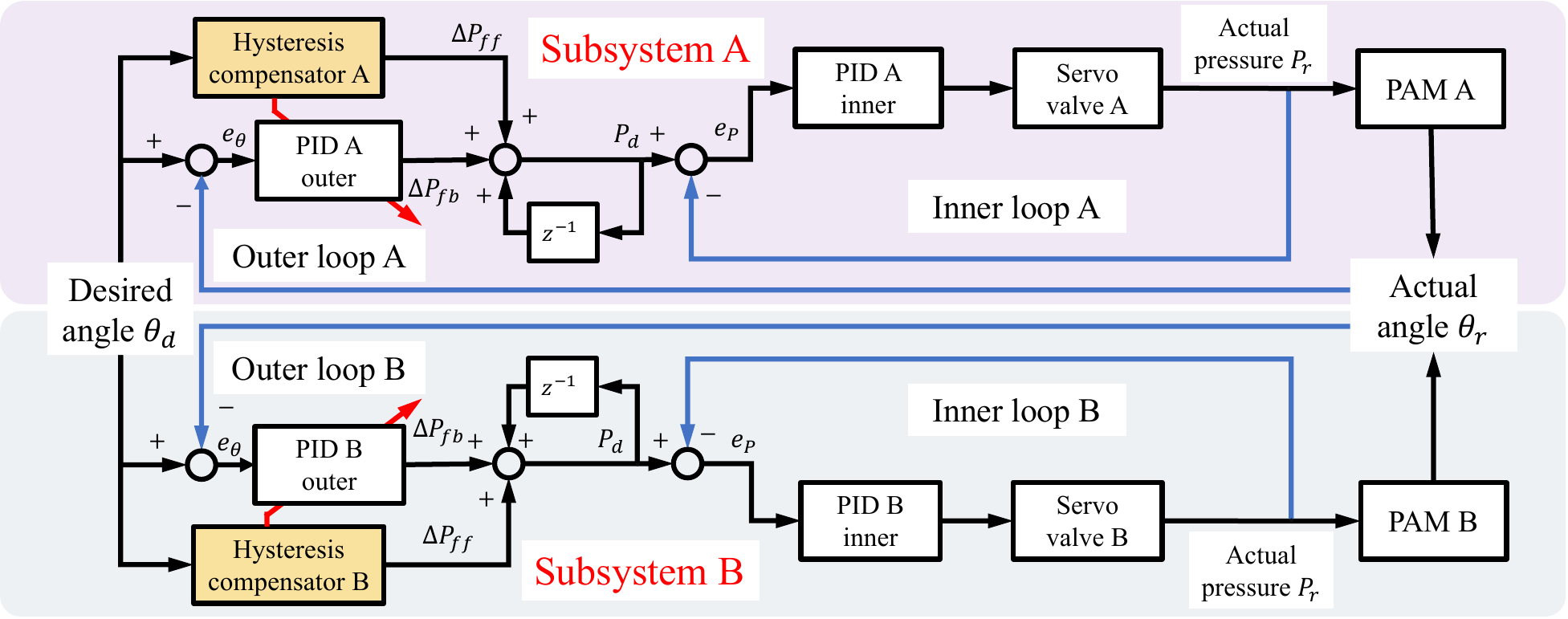}
    \caption{Schematic of the control system with two subsystems independently governing two PAMs.}
    \label{fig:control system}
\end{figure*}

We design the control system in a discrete-time framework, utilizing $k$ to signify the current time step and $k-i, (i=1,2,...)$ to represent the $i$-th time step preceding the current. The architecture of control system comprises two parallel subsystems, each embodying a cascaded structure with an outer loop dedicated to position control and an inner loop for pressure control. Both loops employ PID controllers, with each subsystem regulating a singular PAM of the actuator. This arrangement is depicted in Fig. \ref{fig:control system}, where black lines delineate forward paths from the input reference value to the output actual value, blue lines illustrate the feedback paths, and red lines display the adjustments delivered to the dynamic PID controller. An adaptive compensator is incorporated within each subsystem to account for hysteresis affects. Besides contributing to the adjustments in the pressure reference for the inner loop $P_d$, the compensator also modulates the gain values of the outer loop PID controller. The pressure reference for the inner loops within both subsystems are computed as:
\begin{equation}
P_d(k) =  P_d(k-1) + \Delta P_{fb}(k) + \Delta P_{ff}(k),
\label{pressure_ref}
\end{equation}
where $\Delta P_{fb}$ and $\Delta P_{ff}$ are the adjustments from the outer loop PID controller and the hysteresis compensator, respectively. Given the symmetry in system's structure, $A$ and $B$ to specify different subsystems are omitted herein for brevity.

\subsection{Design of the Adaptive Hysteresis Compensator}

\begin{figure}[tb]
    \centering
\includegraphics[width=\linewidth]{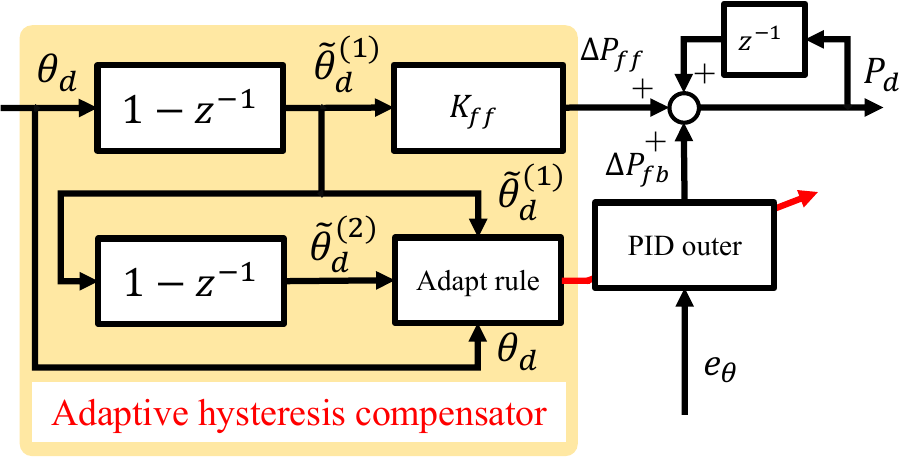}
    \caption{Architecture of the adaptive hysteresis compensator.}
    \label{fig:compensator structure}
\end{figure}

Fig. \ref{fig:compensator structure} illustrates the proposed adaptive hysteresis compensator, encompassing an adaptive program alongside a feedforward differential (D) controller. In this configuration, $\widetilde\theta_d^{(i)}, (i \geq 1)$ denotes the $i$-th order pseudo-differential of the reference bending angle $\theta_d$. The feedforward D gain, $K_{ff}$, operating as:
\begin{equation}
\Delta P_{ff}(k) = K_{ff} \widetilde\theta_d^{(1)}(k),
\label{P_ff}
\end{equation}
aims to mitigate the intrinsic response delay within feedback control and the deformable soft actuator \cite{chen2019fundamental}. The computed $\Delta P_{ff}$ is thereafter merged with the pressure adjustment $\Delta P_{fb}$ from the outer loop PID controller, jointly serving as an adjustment to the inner loop's reference, as shown in Eq.~(\ref{pressure_ref}).

Incorporating the feedforward factor proves notably advantageous when the desired bending undergoes significant variation and exhibits correspondingly high differential values. Contrarily, the adaptive program is expected to operate when the reference changes slightly, where the feedforward component exerts limited effort due to low differential values. The adaptive program processes the reference signal along with its first and second pseudo-differential values, subsequently delivering adjustments to the PID controller within the outer position loop, dynamically altering its proportional (P) gain.

To facilitate a streamlined tuning process of PID settings, a regimen was devised where the variations in internal pressure adjustments within the antagonistic PAMs are designed to reflect each other inversely. Consequently, PID gains in the outer loops of both subsystems and the feedforward differential gains are assigned opposite signs. Concurrently, parameter configurations of the adaptive program, which will be elaborated in the ensuing subsection, are identically set across different subsystems. Additionally, the same PID gains are allocated for inner loops across different subsystems to ascertain consistent performance in pressure control.

\subsection{Adaptive Law of Dynamic Proportional Gain}

Prior to expounding on the adaptive rule depicted in Fig. \ref{fig:compensator structure}, it is imperative to elucidate several pivotal facets pertinent to motion tracking control employing pneumatic bending actuators. (1) Hysteresis-induced delay in actuator response primarily surfaces during state transitions between pressurization and depressurization, typically aligning with turnings in the target motion trajectory where the velocity direction alters. (2) Given any existing target's continuous movement, a deceleration-hold-acceleration sequence is inherent to any directional change in the tracked motion. (3) In the absence of rapid variation in the tracked motion, the reference can be deemed in a quasi-static state, wherein augmenting the proportional gain of conventional PID controllers can enhance tracking performance \cite{LI2018328}. However, an excessive gain increase might instigate instability during large-scale reference changes, as dynamic motion tracking typically demands more stringent gain margin requirements than static state following \cite{ang2005pid}. (4) Given the frequency invariance of the hysteresis property \cite{visintin2013differential}, a viable mitigation strategy involves altering the control signal drastically, thus compelling the system to pass hysteresis dead-zones quickly.

Aiming for wide application, our adaptive program is architected to process the reference signal in real-time. The program employs velocity and acceleration of the reference motion to predict its variation and accordingly modulate the dynamic proportional gain. Given the nonlinear control system's complexity, deriving a precise one-to-one relationship between the dynamic PID gain and the reference signal is scarcely feasible. Moreover, an exact linear relationship could lead to significant dynamic gain fluctuations due to oscillations in real-time discrete pseudo-differential values resulting from unequal sampling intervals, which may subsequently induce system instability. Hence, we propose the adaptive law premised on an accumulative operation. Initially, given a negative acceleration of the reference angle $\widetilde\theta_d^{(2)}(k)$, the dynamic proportional gain adjustment $\Delta K_P(k)$ is expressed as:
\begin{equation}
\Delta  K_P(k) = M_1 \frac{\theta_d(k)}{a_1}\frac{b_1}{b_1+|\widetilde\theta_d^{(1)}(k)|} \frac{|\widetilde\theta_d^{(2)}(k)|}{c_1+|\widetilde\theta_d^{(2)}(k)|}D(k).
\label{KP_1}
\end{equation}
Contrarily, for positive $\widetilde\theta_d^{(2)}(k)$, $\Delta K_P(k)$ is determined by:
\begin{equation}
\Delta  K_P(k) = M_2 \frac{\Theta-\theta_d(k)}{a_2}\frac{b_2}{b_2+|\widetilde\theta_d^{(1)}(k)|} \frac{|\widetilde\theta_d^{(2)}(k)|}{c_2+|\widetilde\theta_d^{(2)}(k)|}D(k).
\label{KP_2}
\end{equation}
In other circumstances, $\Delta K_P(k)$ defaults to 0.

In the aforementioned definitions, $\theta_d(k)$ denotes the reference bending angle, $\Theta$ signifies its predetermined variation range. The term $a$ models the correlation between the width of hysteresis dead-zones and current deformation level, as observed in Fig. \ref{fig:hysteresis}. Term $b$ ensures a smooth variation in the proportional gain, limiting gain augmentation during drastic reference changes while facilitating rapid gain increase as the reference variation decelerates. It influences the balance between the smoothness of the overall gain variations and the disparity in change rates of dynamic gain during periods of fast and slow reference variation. Term $c$, the acceleration coefficient, ensures adjustable amplification in dynamic gain with the angular acceleration of reference bending motion. Augmenting $c$ prevents an excessive increase in the accumulatively adjusted gain during slow reference turnings and ensures a sufficient increase in a short period when the reference experiences a quick direction alteration. By tuning $a$, $b$, and $c$ values, the system can exhibit adjustable anti-hysteresis performance, with the asymmetry in hysteresis loops being addressable by assigning different values to parameters with different footnotes (i.e., 1 or 2 in Eq.~(\ref{KP_1}) and Eq.~(\ref{KP_2})). $D(k)$ is a direction changer defined by :
\begin{equation}
D(k) = -\left(1 + \frac{1 + h(k)}{2} \cdot \mu\right) h(k),
\label{D(t)}
\end{equation}
where $ h(k) = \text{sgn}(\widetilde\theta_d^{(1)}(k) \widetilde\theta_d^{(2)}(k))$ is a sign function, and $\mu$ is an augment operator used to impose a forced decreasing tendency to the dynamic proportional gain for preventing sustained high gain values that may cause system instability. Meanwhile, to uphold a baseline of the dynamic proportional gain as its initial value $K_P(0)$, a cutoff operation:
\begin{equation}
K_P(k)=\max\{K_P(0), K_P(k-1) + \Delta K_P(k)\}
\end{equation}
is performed with each update. $M$ ensures formal dimensional consistency on both sides of equations. By redefining parameters $M_i^* = M_i \frac{b_i}{a_i}, (i=1, 2)$, Eq. (\ref{KP_1}) can be rephrased as :
\begin{equation}
\Delta K_P(k) = M^*_1 \frac{\theta_d(k)|\widetilde\theta_d^{(2)}(k)|}{(b_1+|\widetilde\theta_d^{(1)}(k)|)(c_1+|\widetilde\theta_d^{(2)}(k)|)}D(k),
\end{equation}
and Eq. (\ref{KP_2}) can be expressed by: 
\begin{equation}
\Delta K_P(k) = M^*_2 \frac{(\Theta-\theta_d(k))|\widetilde\theta_d^{(2)}(k)|}{(b_2+|\widetilde\theta_d^{(1)}(k)|)(c_2+|\widetilde\theta_d^{(2)}(k)|)}D(k).
\end{equation}

\section{Experimental Validation}

\subsection{Experimental Setup}

Efficacy of the proposed method is ascertained through a series of experiments, with the performance disparities between our new approach and the standard PID control being delineated through the tracking of three different references. Comparative analyses among traditional PID (PID), PID augmented with the feedforward differential element in the hysteresis compensator (PID+FF), and PID with the entire function of hysteresis compensator (PID+AF) are conducted by tracking sinusoidal references of varying amplitudes and frequencies. The comparative results are tabulated for clarity.

The centroid of the diverse sinusoidal references is fixed at 30°. In the frequency evaluation, the amplitude is maintained at 20°, rendering a reference bending range from 10° to 50°, with periods of 10s, 8s, 6s, 4s, and 2s. In the amplitude analysis, the period is fixed at 8s, while amplitudes of 10°, 15°, 25°, and 30° are tested, given that a 20° amplitude is employed in the frequency test. Owing to the challenges posed in employing either the model-based nonlinear control algorithms or play-operator-based hysteresis compensation techniques for controlling this multi-DoF actuator, comparisons with these methods are omitted in this work.

Due to its accumulative mechanism, the dynamic gain value in our proposed scheme is contingent on historical reference signals in a short horizon. As a result, the demonstrative experiment is designed wherein the reference signal is cyclically applied thrice, with the central application taken as the experimental result to obviate the performance variance induced by initial conditions. In amplitude and frequency evaluations, the central five of seven sinusoidal reference periods are analyzed for the same reason. 

Each experiment is repeated five times, and obtained data are processed independently with the mean values depicted in figures and tables. Considering environmental uncertainties, we use solid lines to represent the average result across different experimental runs and use shaded regions to denote the variations between the maximum and minimum values. A sampling frequency of 500 Hz is employed for capturing reference signals, actual bending angles, PAM pressures, and updating the proportional gain during the experiments. Via trial and error, parameters within the control system are ascertained: the gains for PID controllers in cascade loops, the feedforward D gain in the hysteresis compensator, and all coefficients utilized in the adaptive program. Identified settings are tabulated in Table \ref{gain_PID} and \ref{param_adapt}. To economize on space, only the PID gains of subsystem A in Fig.~\ref{fig:control system} are tabulated, with the gain values for subsystem B being inferable from the system design description in the preceding section.

\begin{table}[htb]
\centering
\caption{PID gains in subsystem A.}
\label{gain_PID}
\renewcommand{\arraystretch}{0.8}
\begin{tabular}{cp{1.5cm}p{1.5cm}}
\toprule
Parameter & Value & Unit\\
\midrule
Inner - P & $4.0\times10^{-2}$ & V/kPa\\ 
Inner - I & $2.0\times10^{-6}$ & V/kPa$\cdot$ s\\ 
Inner - D & 0 & V$\cdot$ s/kPa \\ 
Outer - P & $8.0\times10^{-2}$ & kPa/deg\\ 
Outer - I & $2.0\times10^{-5}$ & kPa/deg $\cdot$ s\\
Outer - D & 0 & kPa $\cdot$ s/deg\\ 
Feedforward - D & $1.0\times10^{-2}$ & kPa $\cdot$ s/deg\\ 
\bottomrule
\end{tabular}
\end{table} 

\begin{table}[htb]
\centering
\caption{Parameters in the adaptive rule.}
\label{param_adapt}
\renewcommand{\arraystretch}{0.8}
\begin{tabular}{cp{1.8cm}p{1.8cm}}
\toprule
Parameter & Value & Unit\\
\midrule
$M^*_1$ & $6.0\times10^{-2}$ & kPa/deg $\cdot$ s\\ 
$b_1$ & $6.0$ & deg/s\\
$c_1$ & $3.2\times10^{4}$ & deg/$\text{s}^2$\\
$M^*_2$ & $9.6\times10^{-2}$ & kPa/deg $\cdot$ s\\ 
$b_2$ & $6.0$ & deg/s\\
$c_2$ & $5.0\times10^{4}$ & deg/$\text{s}^2$\\
$\Theta$ & $60$ & deg\\ 
$\mu$ & $0.6$ & /\\ 
\bottomrule
\end{tabular}
\end{table}

\subsection{Experimental Results and Discussion}

\begin{figure}[tb]
    \centering
\includegraphics[width=\linewidth]{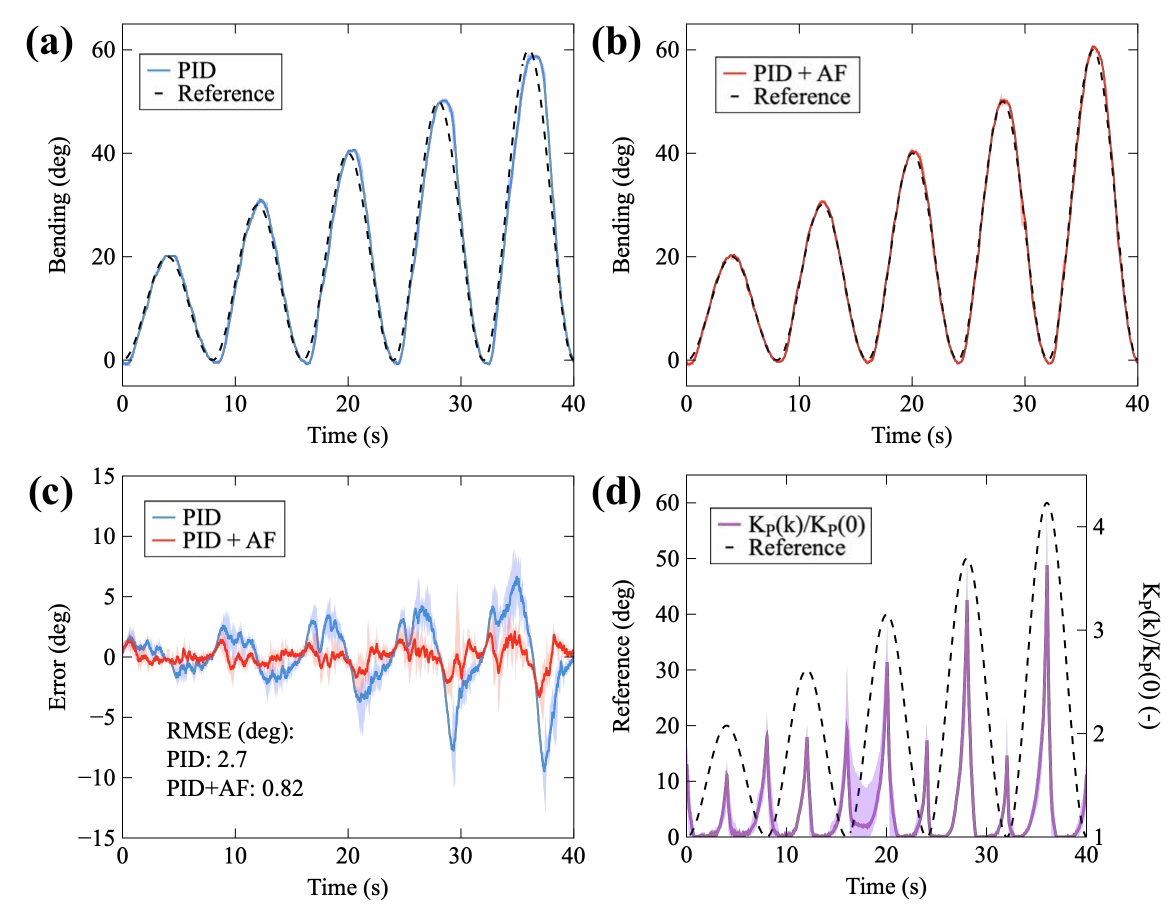}
    \caption{Tracking result 1: (a) PID, (b) Proposed \textit{adaptforward}, (c) Tracking errors, and (d) Dynamic gain amplification.}
    \label{fig:trajectory1}
\end{figure}

\begin{figure}[tb]
    \centering
\includegraphics[width=\linewidth]{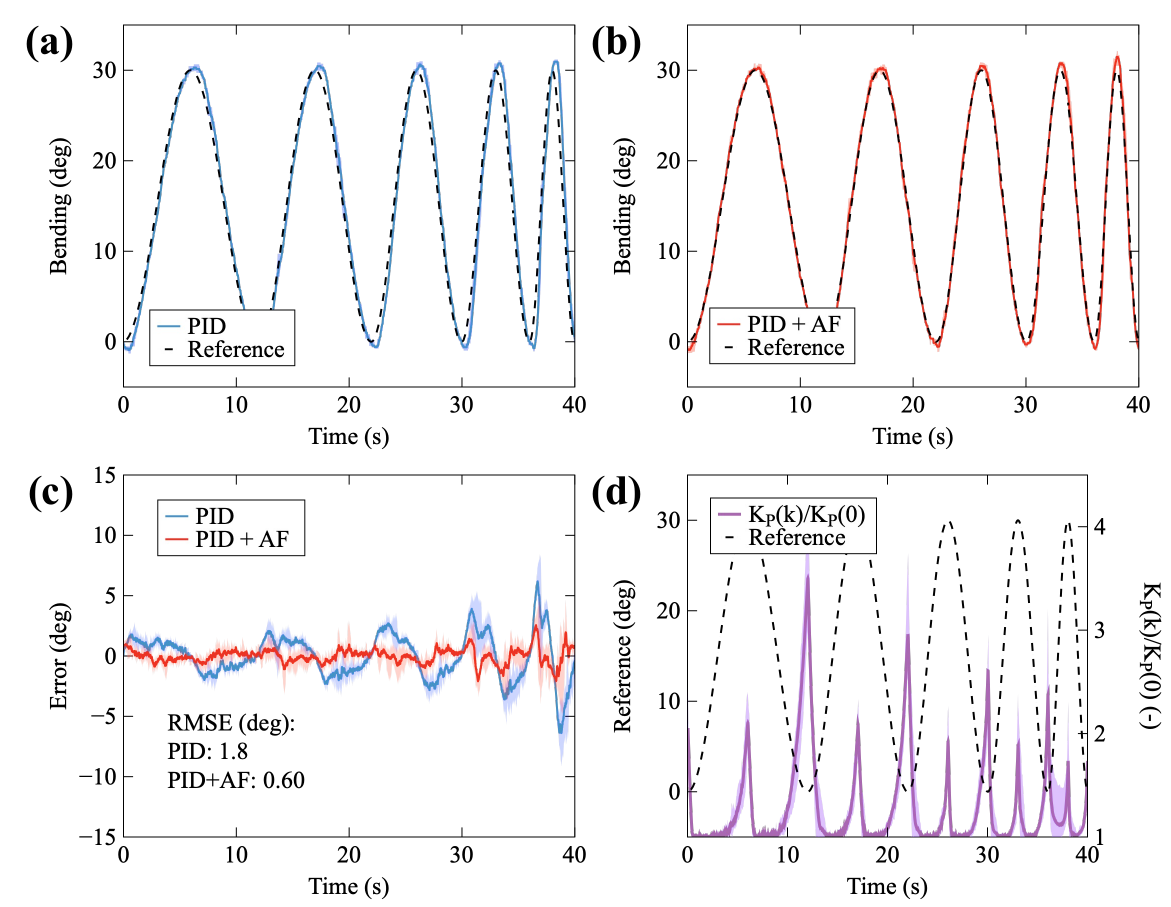}
    \caption{Tracking result 2: (a) PID, (b) Proposed \textit{adaptforward}, (c) Tracking errors, and (d) Dynamic gain amplification.}
    \label{fig:trajectory2}
\end{figure}

\begin{figure}[tb]
    \centering
\includegraphics[width=\linewidth]{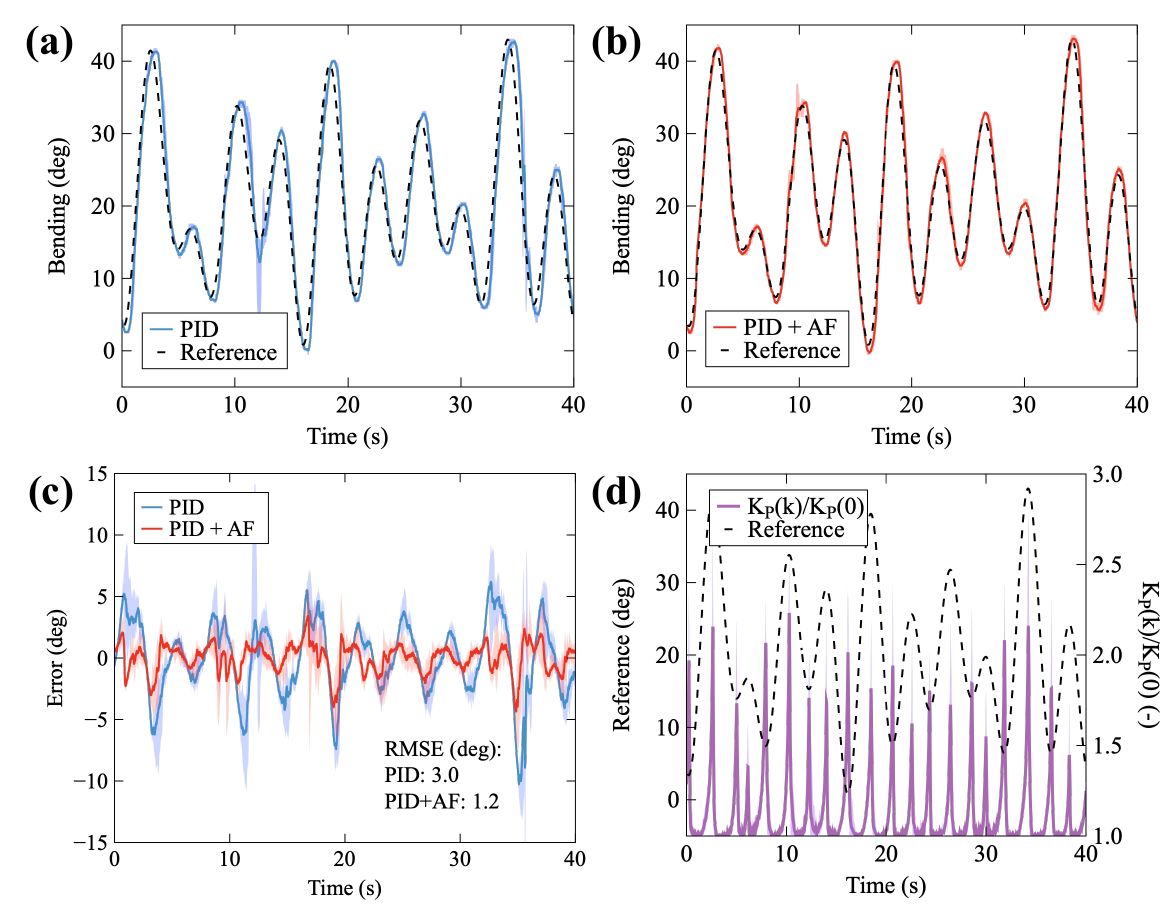}
    \caption{Tracking result 3: (a) PID, (b) Proposed \textit{adaptforward}, (c) Tracking errors, and (d) Dynamic gain amplification.}
    \label{fig:trajectory3}
\end{figure}

Experimental outcomes of tracking three demonstrative references are depicted in Fig. \ref{fig:trajectory1}, \ref{fig:trajectory2}, and \ref{fig:trajectory3}. It is observed that the performance of the traditional PID control considerably deteriorates with the increasing amplitude and frequency of the reference signal, particularly at turning points. Conversely, the proposed PID+AF method bolsters the system's tracking accuracy, ensuring close adherence to the target across various amplitudes and frequencies. The delay in response, attributable to hysteresis, is perceptible in the error patterns, where the errors of PID control exhibit regular fluctuations with changes in the reference signal. In contrast, these fluctuations are markedly subdued in the PID+AF control.

It is observable from experimental results that increases and peaks in amplification ratio of the dynamic gain $K_p(k)/K_p(0)$ coincide with locations where the reference trajectory decelerates to change its direction. The affects of unequal sampling intervals on discrete pseudo-differential operations result in fluctuations in the dynamic gain value, as evidenced by the wide shadow ranges in Fig. \ref{fig:trajectory1} (d) and \ref{fig:trajectory2} (d). However, the adaptive gain demonstrates a steady decline to low values as the reference exits its turning region, thanks to the imposed decreasing tendency in Eq.~(5).

In the case of the compound signal with irregular and frequent changes, fluctuations in response are discernible in both PID and PID+AF results, as shown in Fig. \ref{fig:trajectory3}. Nevertheless, the system maintains stable throughout the repeated experiments. A comparison between the shadow zones reveals that the PID control, despite employing a more conservative constant gain, exhibits more pronounced fluctuations and overshoots, as opposed to the more stable performance exhibited by the proposed PID+AF method.

The performance evaluation of diverse control methodologies across varying reference signals is encapsulated in Table \ref{amp_table} and \ref{freq_table}. A notable degradation in tracking accuracy is observed with escalating frequency and amplitude of the reference for all control approaches. The PID controller exhibits the poorest performance across all experimental scenarios. Incorporating the feedforward differential element (PID+FF) improves the performance of PID, and a further enhancement is realized through deploying the proposed adaptive scheme (PID+AF). Compared to the performance disparity between the PID control and our proposed PID+AF method, the PID+AF delineates a less pronounced enhancement relative to the PID+FF. This occurrence is attributed to the pivotal role the feedforward differential component plays in elevating the performance of PID control, and the adaptive facet is conceived as a compensatory adjunct to the feedforward differential component, particularly at the inflection junctures, where differential values of the reference descend.

\begin{table*}[htb]
\centering
\caption{Performance metrics for tracking references of different amplitudes, with results formatted as PID+AF (PID, PID+FF).}
\label{amp_table}
\renewcommand{\arraystretch}{0.85}
\begin{tabular}{c p{2.5cm}p{2.5cm}p{2.5cm} p{2.5cm} p{2.5cm}}
\toprule
\textbf{Amplitude [deg]} & \textbf{MAE [deg]} & \textbf{RMSE [deg]} & \textbf{$e_\text{max}$ [deg]} & \textbf{$e_\text{min}$ [deg]} & \textbf{Var [$\text{deg}^2$]} \\
\midrule
10 & 0.42 (1.3, 0.56) & 0.51 (1.5, 0.83) & 1.6 (3.3, 2.6) & -1.7 (-4.0, -2.8) & 0.26 (2.3, 0.73)\\ 
15 & 0.53 (1.9, 0.78) & 0.66 (2.3, 0.94) & 2.0 (4.5, 2.8) & -2.3 (-5.9, -2.9) & 0.44 (5.3, 0.88)\\ 
25 & 0.88 (2.5, 1.4) & 1.2 (3.1, 1.7) & 3.8 (6.6, 4.1) & -4.3 (-8.3, -5.7) & 1.5 (9.4, 2.9)\\ 
30 & 1.2 (3.4, 1.8) & 1.7 (4.2, 2.4) & 4.4 (8.4, 6.1) & -6.1 (-11, -9.1) & 3.0 (17, 5.8)\\  
\bottomrule
\end{tabular}
\end{table*}

\begin{table*}[htb]
\centering
\caption{Performance metrics for tracking references of different frequencies, with results formatted as PID+AF (PID, PID+FF).}
\label{freq_table}
\renewcommand{\arraystretch}{0.85}
\begin{tabular}{c p{2.5cm}p{2.5cm}p{2.5cm} p{2.5cm} p{2.5cm}}
\toprule
\textbf{Period [sec]} & \textbf{MAE [deg]} & \textbf{RMSE [deg]} & \textbf{$e_\text{max}$ [deg]} & \textbf{$e_\text{min}$ [deg]} & \textbf{Var [$\text{deg}^2$]} \\
\midrule
10 & 0.56 (1.6, 0.76) & 0.71 (1.9, 1.0) & 2.2 (3.9, 2.7) & -2.4 (-5.1, -3.1) & 0.50 (3.5, 1.1)\\ 
8 & 0.72 (1.9, 1.0) & 0.97 (2.3, 1.5) & 2.8 (5.6, 4.1) & -3.1 (-6.6, -5.0) & 0.94 (5.4, 2.1)\\ 
6 & 0.86 (2.4, 1.7) & 1.3 (3.0, 1.9) & 4.5 (6.8, 5.1) & -4.6 (-8.4, -6.4) & 1.6 (9.0, 3.6)\\ 
4 & 1.5 (4.6, 1.9) & 2.2 (5.4, 2.7) & 5.7 (11, 7.4) & -6.5 (-13, -9.4) & 4.9 (30, 7.6)\\  
2 & 3.9 (9.1, 5.4) & 5.1 (12, 7.2) & 13 (24, 16) & -11 (-22, -17) & 26 (130, 51)\\
\bottomrule
\end{tabular}
\end{table*}

\begin{figure}[tb]
    \centering
\includegraphics[width=\linewidth]{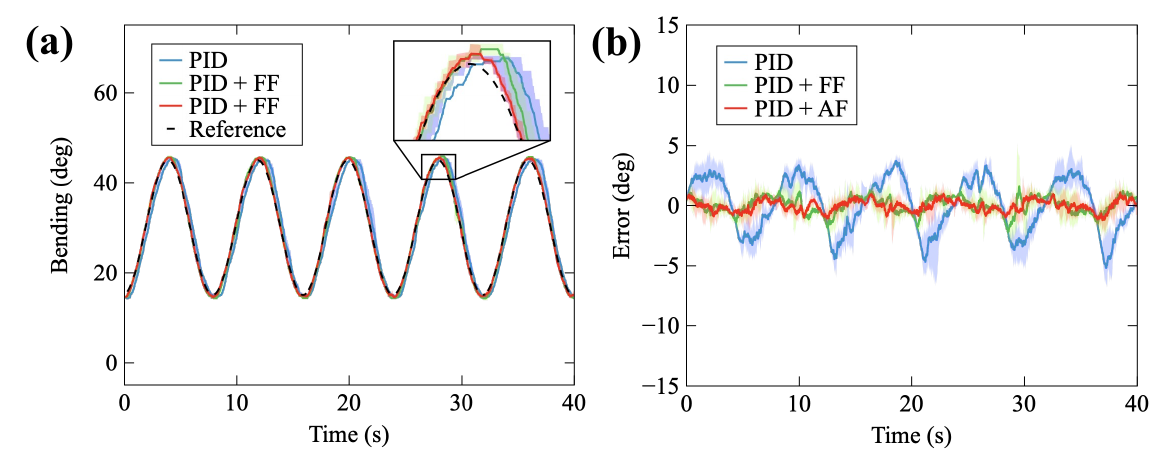}
    \caption{Tracking results of the reference with an amplitude of 15°.}
    \label{fig:amp_1}
\end{figure}

\begin{figure}[tb]
    \centering
\includegraphics[width=\linewidth]{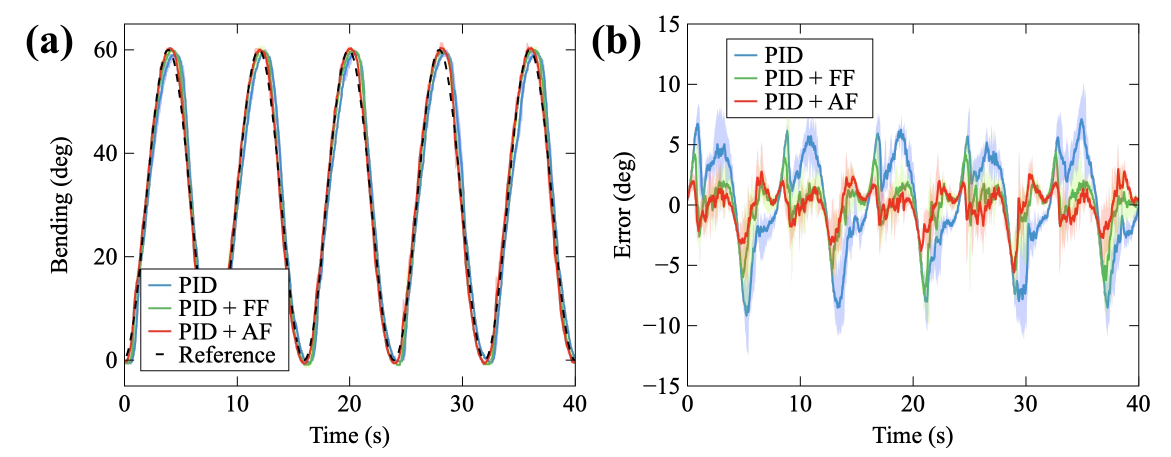}
    \caption{Tracking results of the reference with an amplitude of 30°.}
    \label{fig:amp_2}
\end{figure}

Fig. \ref{fig:amp_1} and \ref{fig:amp_2} delineate the tracking for signals with a low amplitude (15°) and a high amplitude (30°), respectively. For lower amplitudes, PID control alone manifests notable response delays at trajectory turning points. As amplitude intensifies, discrepancies between actual and reference signals become apparent. Both PID+FF and PID+AF ameliorate this discrepancy, attributing to the rapid response brought by the feedforward D element. PID+AF exhibits the minimal shifts in response at turning points, indicating its superior responsiveness to reference alterations. The efficacy of PID+FF dwindles at trajectory turning zones due to the correspondingly low differential values, resulting in suboptimal tracking at these junctures. Conversely, the augmented feedback gains of PID+AF at such regions counteract this performance degradation, engendering enhanced tracking performance.

\begin{figure}[tb]
    \centering
\includegraphics[width=\linewidth]{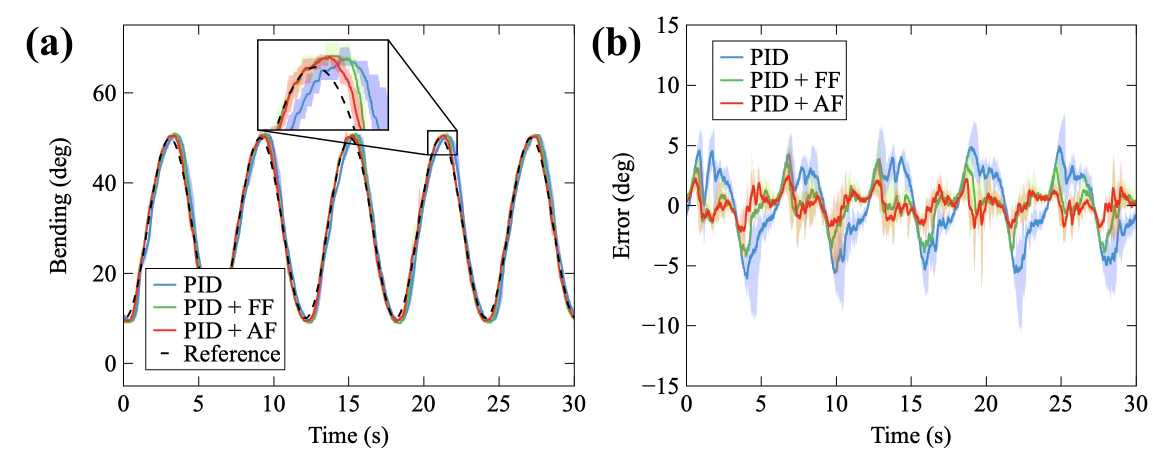}
    \caption{Tracking results of the reference with a period of 6s.}
    \label{fig:freq_1}
\end{figure}

\begin{figure}[tb]
    \centering
\includegraphics[width=\linewidth]{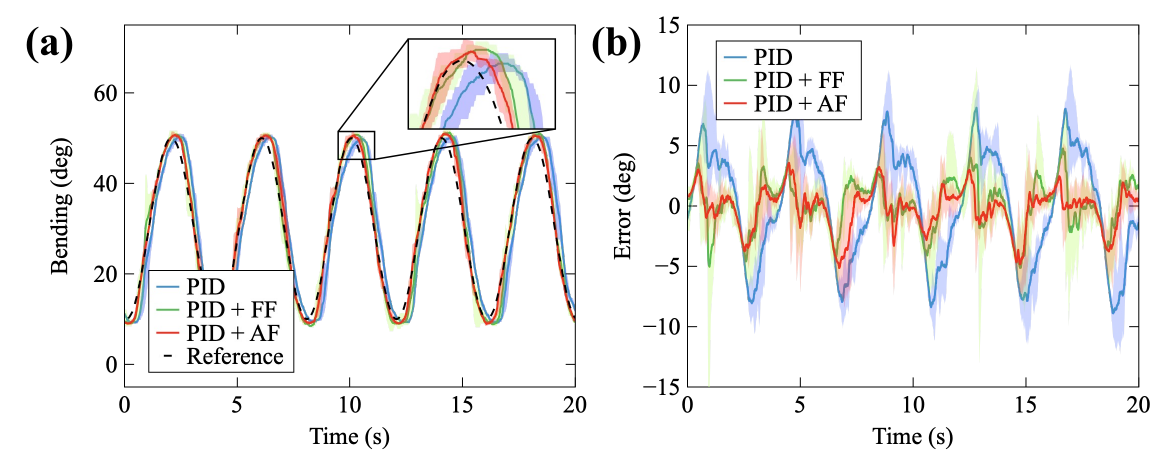}
    \caption{Tracking results of the reference with a period of 4s.}
    \label{fig:freq_2}
\end{figure}

Figures \ref{fig:freq_1} and \ref{fig:freq_2} depict the results for tracking reference signals with a long (6s) and a short period (4s), respectively. Analogous to amplitude analysis, PID control manifests a notable lag at lower frequencies. The PID+AF method's improvement becomes less prominent as frequency elevates, as depicted in Fig. \ref{fig:freq_2}, owing to the predominant error source transitioning from hysteresis to the incompatibility between PID settings and high-frequency signals, which is evidenced by the substantial fluctuation ranges in the PID's response. Although its superior performance is underscored in Table \ref{freq_table}, the proposed PID+AF fails to fully counteract this incompatibility between feedback settings and reference signal characteristics.

\section{Conclusion}

We achieved precise bending trajectory tracking control on a dual-PAM bending actuator paired with a newly proposed adaptive control scheme. The bending actuator comprises two independently controlled PAMs, separated by a central metal plate, with its bending motion hinging on the continuum deformation of both PAMs and the metal plate. The hysteresis characteristics of this pneumatic actuator were assessed through experiments, which reveals the interrelation between the hysteresis dead-zone widths and the deformation levels. 

A control system was introduced for precise bending motion manipulation by amalgamating two cascaded PID loops with newly proposed adaptive hysteresis compensators. The compensator contains an adaptive program and a feedforward differential controller. The adaptive program facilitates dynamic adjustments in the proportional gain of the PID controller for precise bending manipulation under hysteresis affects. The proposed control method, aimed at curtailing hysteresis-induced tracking errors, was experimentally validated on a series of reference signals. 

Our proposed approach is simplistic yet potent in controlling actuators manifesting hysteresis properties, expanding the vista for soft actuator applications in robotics and automation.

\end{document}